\pdfoutput=1
\documentclass[11pt]{article}

\usepackage[]{acl}

\usepackage{times}
\usepackage{latexsym}

\usepackage[T1]{fontenc}
\usepackage[utf8]{inputenc}

\usepackage{microtype}

\usepackage[ruled,vlined,linesnumbered]{algorithm2e}
\usepackage{graphicx}
\usepackage{wrapfig}
\usepackage{amsmath,bm,bbm,upgreek}
\usepackage{amsthm}
\usepackage{xcolor}
\usepackage{caption}
\usepackage{subcaption}
\usepackage{floatrow}
\usepackage{url}
\usepackage{booktabs}
\usepackage{multirow}
\usepackage{wrapfig}
\usepackage{floatflt}
\usepackage{lipsum}
\usepackage{makecell}
\usepackage{mathtools}
\usepackage{dblfloatfix}
\usepackage{xspace}
\usepackage{caption}
\usepackage{subcaption}
\usepackage{floatrow}
\usepackage{bbding}
\usepackage{adjustbox}
\newfloatcommand{capbtabbox}{table}[][\FBwidth]

\usepackage{mdwlist}

\usepackage{cleveref}
\crefname{section}{§}{§§}
\Crefname{section}{§}{§§}

\newif\ifshowcomments
\showcommentstrue
\definecolor{maroon}{rgb}{0.45, 0, 0}

\definecolor{repgreen}{RGB}{0, 100, 0}
\definecolor{insblue}{RGB}{30, 144, 255}
\definecolor{swapyell}{RGB}{184, 134, 11}
\definecolor{delred}{RGB}{255, 99, 71}

\usepackage{easyReview}
\newcommand{\method}{\textsc{SEScore}\xspace}

\title{Not All Errors Are Equal: Learning Text Generation Metrics using Stratified Error Synthesis}

\author{
  Wenda Xu,
  Yilin Tuan,
  Yujie Lu,
  Michael Saxon, 
  \\
  \textbf{Lei Li}, 
  \textbf{William Yang Wang}
  \\
  UC Santa Barbara
  \\
  {\{wendaxu,ytuan,yujielu, mssaxon, leili, william\}@cs.ucsb.edu} \\
}

\begin{document}
\maketitle
\begin{abstract}
Is it possible to build a general and automatic natural language generation (NLG) evaluation metric? 
Existing learned metrics either perform unsatisfactorily or are restricted to tasks where large human rating data is already available. We introduce \method, a model-based metric that is highly correlated with human judgements without requiring human annotation, by utilizing a novel, iterative error synthesis and severity scoring pipeline. This pipeline applies a series of plausible errors to raw text and 
assigns severity labels by simulating human judgements with entailment. 
We evaluate \method against existing metrics 
by comparing how their scores correlate with human ratings.
\method outperforms all prior unsupervised metrics on multiple diverse NLG tasks including machine translation, image captioning, and WebNLG text generation.
For WMT 20/21 En-De and Zh-En, \method improve the average Kendall correlation with human judgement from 0.154 to 0.195.
\method even achieves comparable performance to the best supervised metric COMET, 
despite receiving no human-annotated training data. \footnote{Code and data are available at \url{https://github.com/xu1998hz/SEScore}}
\end{abstract}

\section{Introduction}
\label{sec:intro}
Text generation tasks such as translation and image captioning have seen considerable progress in the past few years \cite{Chen2015MicrosoftCC, 625af6f84b724fcda0a8bf6026cc922f}. %However, using an automatic evaluation to accurately access text generation performance still remains as a core obstacle. 
%However, it remains a challenge to automatically and precisely evaluate generated text quality. 
However, precisely and automatically evaluating generated text quality remains a challenge.
% \yj{I don't think this obstacle sentence is appropriate here, since you are illustrating some previous work in the following sentence}
%including n-gram based approaches 
Long-dominant n-gram-based evaluation techniques, such as BLEU \cite{papineni2002bleu} and ROUGE \cite{lin-2004-rouge}, are sensitive to surface-level lexical and syntactic variations, and have been repeatedly reported to not well correlate to human judgements \cite{bert-score, xu2021selfsupervised}. 

%% Lei Li breaks the paragraph here.

Multiple \emph{learned metrics} have been proposed to better approximate human judgements.
These metrics can be categorized into  \emph{unsupervised} and \emph{supervised} methods based on whether human ratings are used. 
The former includes PRISM~\cite{thompson-post-2020-automatic}, BERTScore~\cite{bert-score}, BARTScore~\cite{Yuan2021BARTScoreEG}, etc. 
The latter includes BLEURT~\cite{sellam2020bleurt}, COMET~\cite{rei-etal-2020-comet} etc. 

Unsupervised learned metrics are particularly useful as task-specific human annotations of generated text can be expensive or impractical to gather at scale. % (e.g, low resource languages in Machine Translation). 
While these metrics are applicable to a variety of NLG tasks \cite{bert-score, Yuan2021BARTScoreEG}, 
they tend to target a narrow set of aspects such as semantic coverage or faithfulness, and have limited applicability to other aspects, such as fluency and style, that matter to humans \cite{freitag2021experts, saxon-2021-modeling}. While supervised metrics can address different attributes by modeling the conditional distribution of real human opinions, 
%full advantage of available human rating data and fits the distribution tightly to the human rating data. However, this approach can have 
training data for quality assessment is often task- and domain-specific with limited generalizability. 
%domain shifts when adapting to dataset in different domains or even the same dataset with different human labels. 
%\ms{Might be important to make this point when rationalizing what we're doing here: style and fluency are subjective notions, but the fact that readers tend to agree that some style and fluency are more preferable to others is objective.}

%\wenda{We can not say subjective here. Because our approach is trying to prove they are not subjective criteria but can be accessed through our pipeline}

% For example, COMET which trained on WMT 17-19 testing set with crowd source labels can have high human correlation in WMT 20 with crowd sourcers ratings. However, the correlation dramatically drops when testing it on WMT 20 with expert-based MQM labels. Therefore, an ideal learned metric would not be dependent on specific human rating distribution and it can cover diverse error types and is able to be extensible to multiple domains and tasks.  \ww{There's a big issue with your intro is that you spent way too much space (about 80 lines) on motivations and other people's work. Your own work is only described within 116-128, these 20 lines. You need to re-balance the intro.}

We introduce \method, 
a general technique to produce nuanced \emph{reference-based} metrics for automatic text generation evaluation without using human-annotated reference-candidate text pairs.  
%that capture human notions such as fluency and accuracy, \textbf{in any domain}, \textit{without requiring human-annotated training data}.
%In this work, we introduce a \textbf{stratified error synthesis} process for pretraining a \textbf{reference-based} automatic evaluation metric that models human notions like fluency and accuracy \textit{without annotated data}. 
%Reference-based metrics measure the extent of similarities between candidate and reference without requiring source data. 
Our method is motivated by the observation %, during human evaluation, that errors in candidate text can belong to multiple types and are not always critical 
that a diverse set of distinct error types can co-occur in candidate texts, and that human evaluators do not view all errors as equally problematic \cite{freitag2021experts}.
To this end, we develop a \emph{stratified error synthesis} procedure to construct (reference, candidate, score) triples from raw text. 
The candidates contain non-overlapping, plausible simulations of NLG model errors, iteratively applied to the input text. 
At each iteration, a \emph{severity scoring} module isolates individual simulated errors, and
assesses the human-perceived degradation in quality incurred. 
%To accomplish this, we construct a diverse set of error types to simulate the real text generation errors and incrementally inject them into the raw text.  The creation of each error has to take the consideration of all the
%prior errors and posterior synthetic errors will not modify or overwrite the prior ones. %Inspired by expert-based MQM framework \cite{Mariana2014TheMQ}, we assign one score to each error in our error synthetic sentences and final score of each sentence is estimated based on how many errors they contain and each error's severity levels. 
%In general, \method is applicable to quality assessment for any NLG task. 
Our contributions are as follows:
\begin{itemize*}
    \item \method, an approach to train automatic text evaluation metrics without human ratings;
    \item A procedure to synthesize different types of errors in text at varying severity levels;
    \item Experiments showing that \method is effective in a diverse set of NLG tasks including WMT 20/21, WebNLG, and image captioning, and outperforms all previous unsupervised learned metrics. It is even comparable to the best learned metric on WMT 20/21. 
\end{itemize*}

\section{Related Work}
Traditional n-gram matching based ~\citep{papineni2002bleu, banerjee-lavie-2005-meteor} and edit distance based approaches~\citep{Levenshtein1965BinaryCC, snover-etal-2006-study} have proven to be limited in recognizing semantic similarity beyond the lexical level.
Learned metrics \cite{bert-score, sellam2020bleurt, Yuan2021BARTScoreEG} have been proposed to align better with human judgements. We categorize these metrics as either unsupervised or supervised with respect to learning from human-annotated scores.

\paragraph{Unsupervised Metrics} attempt to extract features from large pretrained models.
% Therefore, they do not require any human rating data to form a metric.
Embedding-based metrics (e.g. BERTScore \cite{bert-score} and Moverscore \cite{zhao2019moverscore}) create soft-alignments between reference and hypothesis in the embedding space. However, they are refined in the semantic coverage. Text generation-based metrics \cite{Yuan2021BARTScoreEG}, use conditional probability of the generated sentence to evaluate faithfulness of the candidates. However, \citet{freitag2021experts} points out text generation evaluation can produce errors beyond semantic coverage or faithfulness (e.g. style and fluency errors), which results poor correlations to the human evaluations.  

\paragraph{Supervised Metrics} attempt to learn through limited human-labelled severity annotations.
% to estimate text quality. It has recently proven to be effective in many natural language generation tasks (NLG), such as Machine Translation.
% \citet{rei-etal-2020-comet} uses both ranking and regreesion model to train on the WMT 17-19 testing set \cite{bojar-etal-2017-results, specia-etal-2018-findings, ma-etal-2019-results} with human labels.
% It is the best performed evaluation metric in WMT 2020 shared task \cite{mathur-etal-2020-results}.
% However, human ratings are scare in many tasks in NLG. Since COMET is trained on a small domain of human rating data, it is difficult to extend it to the general domain.
\citet{rei-etal-2020-comet} trained COMET on a small set of domain-specific human ratings; this model has limited extensibility to teh general domain.
% \citet{sellam2020bleurt} is a BERT-based \cite{devlin2019bert} metric.
BLEURT~\citep{sellam2020bleurt} first pretrains on millions of synthetic data and then uses WMT testing data in fine-tuning the model. Unlike our fine-grained stratified error synthesis, the labels on the synthetic data are derived from prior metrics or other tasks,  limiting  the quality and precision of pretraining process.

\section{The \method Approach}
\label{sec:method}

\begin{figure}
    \centering
    \small
    \includegraphics[width=.97\linewidth]{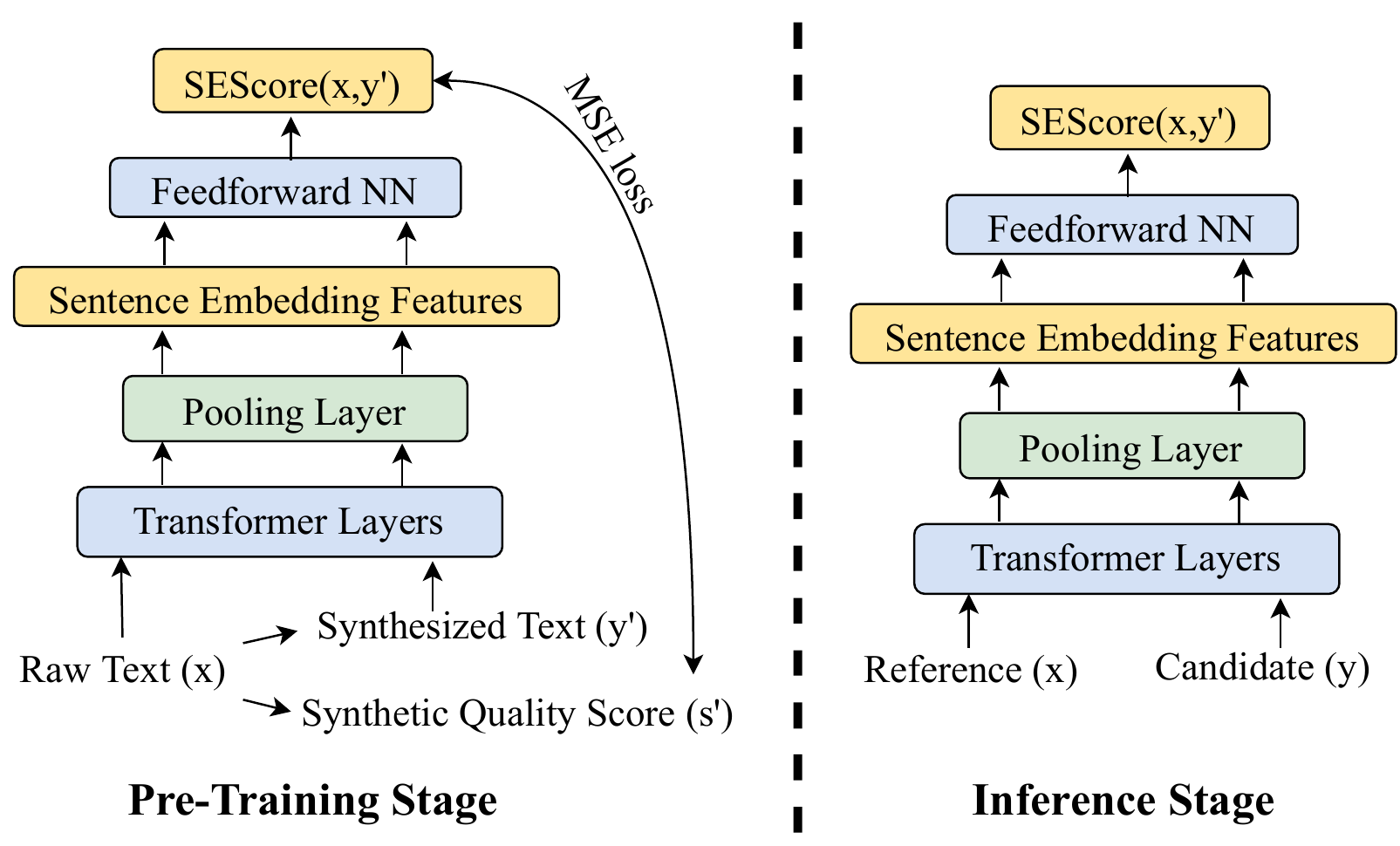}
    \caption{Overview of the Quality Prediction Model.}
    \label{fig:model}
\end{figure}

\begin{figure*}
    \centering
    \includegraphics[width=.98\linewidth]{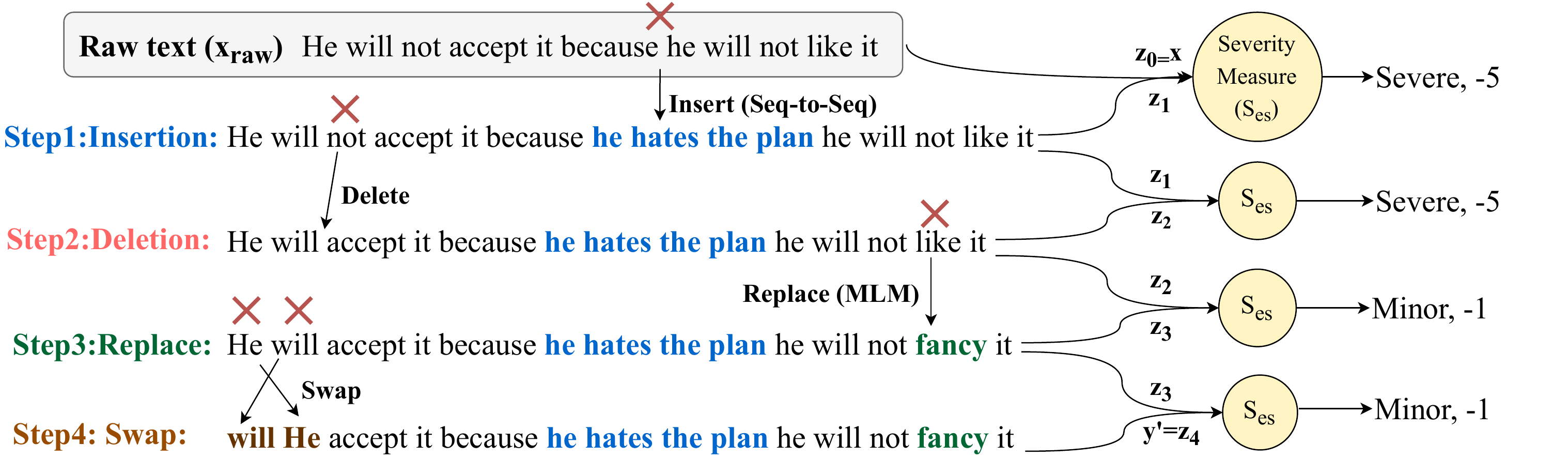}
    \caption{\method: stratified error synthesis and severity scoring Pipeline. {\color{maroon}{\footnotesize\XSolid}} indicates the start index of each error in the previous sentence. Both MLM and Seq-to-seq models can be used to produce inserted or replaced tokens. Each $\mathbf{z}_i$ corresponds to a perturbed sentence. The final synthesized sentence $y'$ has the score $s'=\sum_{i=1}^{4} S_{\text{es}}(\mathbf{z}_{i-1}, \mathbf{z}_{i})=-12$.}
    \label{fig:method}
\end{figure*}

Given a reference text $\pmb{x}$ and a candidate $\pmb{y}$, a metric is expected to output a score $s$. 
Training such a metric model requires triples of reference-candidate-score's. 
However, there are no large-scale human annotated triple data available in many tasks. 
We consider a general setup where large raw text corpus is available.

%Ideally, supervised learned metrics estimate the quality $s$ of a text pair $(x,y)$ using a model $f_\theta$ trained with objective $\text{argmin}_\theta(\mathcal{L}(f_\theta(x,y),s))$, from some corpus of human-scored reference-candidate pairs.
%However, this framework is restricted to problems and domains where large corpora of $\{\langle x, y, s \rangle\}$ triples are available for training. 

%=g_\text{final}(x) %=S_\text{final}(x,g(x))

\method is %built on a pre-trained language model (e.g. BERT) and further 
trained from a pretrained language model (e.g. BERT) on synthetic triples generated from raw text. 
It synthesizes candidate sentences $\pmb{y}'$ to mimic plausible errors by transforming raw input sentences $\pmb{x}$ multiple times. 
At each step, it inserts, deletes, or substitutes a random span of text. 
These errors are non-overlapping.
It assesses the severity of the errors introduced in the transformation. 
This allows us to pretrain quality prediction models %$f_\theta$ 
on corpora containing only raw text samples $\{x\}$, enabling the use of learned quality prediction models in any text generation domain.

% a pretraining technique that overcomes the limitations of supervised learned metrics to domains where such human ratings are unavailable,
% by synthesizing model output-like sentences $y'$ from raw text samples $x$ using a series of $M$ transformations which each can be individually assessed in severity to produce 
% synthetic quality scores $s'$ for training. 

The process of generating $\pmb{y}'$ from $\pmb{x}$, \textbf{stratified error synthesis}, 
is so called for its incremental and multi-category nature; a stochastic perturbation function $G_{\text{es}}$ which randomly %draws one perturbation each time from $E$, where $E=\{e_{repl},e_{ins},e_{del},e_{swap}\}$, 
samples from a set of potential errors 
is recursively applied on $\pmb{x}$ (\cref{eq:eqn1}) $M$ times to produce a sequence of perturbed sentences $\pmb{Z} =\{\mathbf{z}_i\}_{i=1}^{M}$ that interpolate between the raw text $\pmb{x}$ and the final synthetic sentence $\pmb{y}'=\mathbf{z}_M$ (\cref{sec:ses}).
% \yt{change all the $k$ to $M$ here}
% \yt{change eq1, $0<i\leq M,  1<M\leq N=5$} %\textit{stratum}
\begin{align}
    \label{eq:eqn1}
    {\mathbf{z}}_i=\begin{cases}
    \pmb{x}, & \text{if } i=0 \\
    G_{\text{es}}(\mathbf{z}_{i-1}), & \text{0$<i\leq M$}
    \end{cases}
        %& \left\{\right .
\end{align}

The stratum sentence sequence $\mathbf{Z}$ is then used to in the subsequent \textbf{severity scoring step} which uses a pairwise severity scoring function $S_{\text{es}}$ on consecutive pairs and cumulatively yield training labels $s'=\sum_{i=1}^{M} S_{\text{es}}(\mathbf{z}_{i-1}, \mathbf{z}_{i})$ (\cref{sec:sevscore}).
A concrete example is illustrated in \cref{fig:method}. Finally, we train \method's \textbf{quality prediction model}, $f_\theta$ ( \cref{fig:model}) using synthetic $\{\langle \pmb{x}, \pmb{y}', s'\rangle\}$ triples (\cref{sec:qualpred}). 

\begin{table*}
  \label{langtable}
  \centering
  \small
  \resizebox{\textwidth}{!}{
  \begin{tabular}{llll}
    \toprule
    Category &  & MQM Description & Synthesis Procedure in \method\\
    \midrule
    Accuracy & Addition & Text includes information not present reference. & insertion using MLM or seq2seq generation\\
    & Omission & Text is missing content from the reference & Delete a random span of tokens \\
    & Mistranslation & Text does not accurately represent the reference & Replace a random span using maksed or seq2seq generation \\
    Fluency & Punctuation & Incorrect punctuation (for locale or style) & Insertion \& replacement using masked filling, and deletion \\
    & Spelling & Incorrect spelling or capitalization &  
    Insertion, replacement, deletion, and Swap\\
    & Grammar & Problems with grammar & Insertion, replacement, deletion, and Swap \\
    \bottomrule
  \end{tabular}
    }
    \caption{
    Error Categories in MQM and our synthesis procedure. \method generalize the imitate model output errors beyond machine translation.   % \yj{name for the second column?}
    }
  \label{tab:error_type}
\end{table*}

\subsection{Background: Quality Measured by Errors}
Our method is inspired by the multidimensional quality metrics (MQM)~\cite{Mariana2014TheMQ, freitag2021experts}.
MQM is a human evaluation scheme for machine translation. 
It determines the quality of a translation text by manually labeling errors and their severity levels. 
Errors are categorized into multiple types such as accuracy and fluency. Each error type is associated with a severity level -- a penalty of 5 for major error and 1 for minor error. 

In \cref{tab:error_type}, we use two major error categories in MQM framework: accuracy and fluency, to classify and decide our perturbations in $G_{\text{es}}$. There are two main motivations to simulate those errors from the table: 1) they are two major error categories in machine translations; 2) those errors are general and can be extensible to new domains. We use six techniques to simulate errors from the \cref{tab:error_type}:
mask insertion/replacement with maksed language model (MLM)/seq-to-seq (seq-to-seq) language model, and N-gram word drop/swap.

\subsection{Stratified Error Synthesis}\label{sec:ses}

% We follow MQM's typology \cite{freitag2021experts} of major and minor errors that can impact perceived accuracy and fluency as shown in Table \ref{tab:error_type} in choosing the types of errors we simulate.

\citet{tuan2021quality} suggest that multiple errors could co-occur in one segment, so we construct each sentence with up to $M_{\text{max}}$ perturbations ($=5$ in experiments). At each iteration, we randomly draw one perturbation $G_{\text{es}}$ from the set of edit operations, $E={\{e_{ins}, e_{del}, e_{repl}, e_{swap}\}}$ (insertion, deletion, replacement, and swap, respectively).

Our technique is \emph{stratified} so as to enable accurate evaluation of the severity at each step, and prevent subsequent errors from overwriting prior ones.
%, the creation of each error has to take the consideration of all the prior steps and posterior perturbations will not overwrite the prior ones. 
To achieve this, we propose a novel stratified error synthesis algorithm. For an input sentence $\pmb{x}$, with $L$ tokens, we initialize an array $q$ of length $L$, with $q_j = L-j, \forall 1\leq j\leq L$. 
Values indicate the number of tokens after the current token can be modified with the perturbation function, $G_{\text{es}}$. Each $G_{\text{es}}$ will randomly select a start index $j$ from $1$ to $L$ to modify the text. We define an error synthesis table to keep track of the number of candidate tokens can be modified after index $j$. $G_{\text{es}}$ will only be accepted if $q_j$ is greater than the span length of the perturbation. The implementation details of stratified error synthesis algorithm regarding to each edit operation is illustrated in Appendix A \cref{alg:SSE}. All perturbations are recursively applied to the raw text $\pmb{x}$, shown in \cref{eq:eqn1}. 

% \ww{I got very confused with your algorithm, because it is not obvious what order of editing operations you will use, how to create diverse set of errors, and how to choose the position to edit.} \ww{For such key algorithm, I do not think putting that in the appendix is the right thing to do.} 

\paragraph{Synthesize Addition Error by Insertion ($e_{ins}$)}
Given a start index, we add an additional phrase to the raw text in
two ways: a) using a MLM (e.g. BERT and RoBERTa), and b) using a seq-to-seq language model (e.g. mBART). 
% \yj{merge}
For the first approach, we insert a \verb|<mask>| token at the given position of a sentence. Then, we use an MLM to fill the token based on its context. 
We use top-k sampling ($k=4$), to randomly select the filling token.
% \ww{Not clear to readers how to get the ``given position'' from this context, unless they scrolled to the very back to look at your sampling algorithm?}
% \ww{check all typos using my grammarly premium account.}
% Many of those <mask>s are randomly placed within the word.
% Yi-Lin: "a" word
Our primary aim is to introduce semantically close sentences with all three \emph{fluency} errors. 
With the insertion of \verb|<mask>|, we can further synthesize \emph{Addition} errors.
% \yj{merge}
For the second approach, we use a pre-trained seq-to-seq model (e.g. mBART) to generate a phrase given the context text, with variable length. 

% Mask Insertion and Replacement with Roberta\ww{Be consistent with spelling and capitalization. Consult original paper for capitalization.}.

% RoBERTa's mask infilling objective enables the model to predict on the <mask> token at its left and right context.
% \ww{I'm not sure if we should name a specific in-filling model like RoBERTa in this approach section: imagine if 5 years later, no one is using RoBERTa anymore, but your paper can still be relevant, and people can use other in-filling LMs, correct?}
% Yi-Lin: what does "on the <mask> token at its left and right context" mean?

% Yi-Lin: how does insertion and replacement can be applied in the same time?

\paragraph{Synthesize Omission Error by Deletion ($e_{del}$)}
We delete a random span of tokens from a raw text sentence. The span is drawn uniformly within the token indices. The length of the span is drawn from a Poisson distribution ($\lambda_d=1.5$).
Our primary aim is to mimic \emph{Omission} error.
However, depending on the specific words that it drops, this technique can further create \emph{Mistranslation} and all \emph{Fluency} errors. 
% \yj{Initialize Paragraph, if it's not part of the sentence? check through the paper}

\paragraph{Synthesize Phrasal Error by Replacement ($e_{repl}$)}
Sometimes specific terms in a reference sentence are systematically misphrased in generated samples. 
%While comparing difference between the reference and candidate sentence, certain phrases can be erroneous. 
This is difficult to simulate. 
Instead, we use either an MLM or a seq-to-seq model to replace a segment of tokens in the original text.
% \yj{merge}
For the first approach, the replaced span is always a single token, which is first replaced with a \verb|<mask>| token. 
We then use an MLM to fill the blank similar to the insertion operation. 
% \yj{merge}
For the second approach, we use a denoising seq-to-seq model (e.g. mBART) to generate tokens for the mask tags. 
We randomly choose the starting index of the span and draw the span length from a Poisson distribution ($\lambda_d=1.5$).
We use a denoising seq-to-seq model like mBART to synthesize fluent sentences with \emph{Addition} and \emph{Mistranslation} errors. 
% Through the model's rewriting capability, we can dynamically modify sentence length through injected error locations.  

% Different from Roberta's mask infilling objective, MBart predicts a span of tokens for its <mask> position through auto-regressive rewriting. We leverage this functionality by inserting and replacing a span of tokens in the raw text though MBart. We draw a possion distribution with $\lambda=1.5$ to determine the replaced span length. Since MBart is trained as a denoising autoencoder, We expect this rewriting to produce fluent sentences but introducing 'Addition' and 'Mistranslation' errors. Through MBart's rewriting, we can dynamically modify sentence length through injected error locations.  

\paragraph{Synthesize Grammar and Other Errors by Swapping ($e_{swap}$)}
We swap two random words within the span length $\lambda_s$ in the sentence ($\lambda_s=4$). Our primary aim is to generate grammatically incorrect sentences with mismanagement of word orders, such as subject verb disagreement. It further introduces \emph{Spelling} and \emph{Punctuation} errors.

\subsection{Assessing Severity Score}\label{sec:sevscore}
% As suggested by the recent study \cite{freitag2021experts}, different errors in the sentence can have different severity measures, depending on the level that they mislead the readers.
Following \citet{freitag2021experts}, we consider an error \emph{severe} if it alters the core meaning of the sentence. 
Prior study has suggested that sentence entailment is strongly correlated to semantic similarities \cite{khobragade2019machine}.
To capture the change of semantic meaning, we define a bidirectional entailment relation such that, text $a$ entails $b$ and $b$ entails $a$ is equivalent to $a$ is semantically equivalent to $b$.
Therefore, for a given perturbation function $G_{\text{es}}$ on the sentence $\mathbf{z}_{i-1}$, we measure a bidirectional entailment likelihood of $\mathbf{z}_{i-1}$ and $\mathbf{z}_{i}$.
If after applying transformation on $\mathbf{z}_{i-1}$, $\mathbf{z}_{i}$ remains bidirectional entailed to $\mathbf{z}_{i-1}$, we can assume that $G_{\text{es}}$ does not severely alter the semantic meaning of $\mathbf{z}_{i-1}$ and therefore it is a minor error. We define the entailment likelihood, $\rho(a, b)$, as the probability of predicting $a$ entails $b$. The math formulation is illustrated in \cref{eq:thres}. Setting the threshold $\gamma$ to be $0.9$ reaches the highest inter-rater agreement of severity measures using our validation dataset. Following \citet{freitag2021experts}, we assign $-5$ to severe error and $-1$ to minor errors. Therefore, our range of score is $[-25, 0]$. We evaluate severity at each perturbation of the sentence and cumulatively yield training label $s'$ for the final synthesized sentence $\pmb{y}'$, $s'=\sum_{i=1}^{N} S_{\text{es}}(\mathbf{z}_{i-1}, \mathbf{z}_{i})$.
\begin{equation}
\label{eq:thres}
\begin{split}
        & \text{$S_{\text{es}}$}(\mathbf{z}_{i-1}, \mathbf{z}_{i})=\\
        & \begin{cases}
    -1, & \text{if } \rho(\mathbf{z}_{i-1}, \mathbf{z}_{i}) \geq  \gamma \text{ and }\rho(\mathbf{z}_{i}, \mathbf{z}_{i-1}) \geq \gamma\\
    -5, & \text{ otherwise}
        \end{cases}
\end{split}
\end{equation}
% \iffalse
% \begin{align*}
%         & \text{$S_{\text{es}}$}(\mathbf{g}_{i-1}, \mathbf{g}_{i})=\\
%         & \begin{cases}
%     -1, & \text{if } \rho_{\mathbf{g}_{i-1}\rightarrow \mathbf{g}_{i}} \geq  \theta \text{ and }\rho_{\mathbf{g}_{i}\rightarrow \mathbf{g}_{i-1}} \geq \theta\\
%     -5, & \text{ otherwise}
%     \end{cases}
%         %& \left\{\right .
% \end{align*}
% \fi

% We synthesize data from the News Complimentary datasets. For each instance, we create candidate pseudo output sentences using six different noise functions. Each noise is applied iteratively with the maximum iteration to be 5. We draw a Possion distribution ($\lambda=1.5$) to determine the number of noises applied to each instance.  

%For each step, one of the six noise functions will be randomly selected and applied to a random positions of the sentence. After the pseudo output sentence is generated at each step, We use a classification model to determine error's severity. Each error will be given either score of -1 or -5, based on its severity measure. After all the error steps completed on each sentence, we obtain a pseudo sentence quality rating based on all prior severity measures. Based on the prior study \cite{freitag2021experts}, 

\subsection{Quality Prediction Model}\label{sec:qualpred}
In \cref{fig:model}, we fed both raw text $\pmb{x}$ (reference) and synthetic error sentence $\pmb{y}'$ into a pre-trained language model (e.g. BERT or RoBERTa). The resulting word embeddings are average pooled to derive two sentence embeddings. Then we use the approach proposed by RUSE \cite{shimanaka-etal-2018-ruse} to extract the two features: \textit{1)} Element-wise synthesized and reference sentence product. \textit{2)} Element-wise synthesized and reference sentence difference. Following the COMET \cite{rei-etal-2020-comet} implementation, the above  features are concatenated into a single vector and fed into a feed-forward neural network regressor, $f_\theta$. 

However, the key distinction between our model and COMET is that we don't use model source input during training or inference. Therefore our \method can generalize to other text generation tasks, without considering specific source data. The detailed architecture choice can be found in \cref{sec:pretrain_setup}.

\section{Experiments}

% 1. Does \method determine the quality of the generated texts in multiple tasks?
% 2. For MT, can it determine all different models in the same task?
% 3. If Stratified approach improves the evaluation
% 4. What are the effects of different error synthesized functions 
% 5. How the size of data quantity can affect the evaluation performance
% 6. Case study for severity (Prominent examples). Limitations: 

We conduct experiments on three tasks: machine translation, data-to-text and image captioning, to verify the utility and generalizability of  \method. Specifically, we compare \method on WMT 2020 and 2021 test sets in English-to-German (En-De) and Chinese-to-English (Zh-En) with MQM labels~\cite{Mariana2014TheMQ, freitag2021experts}, which consists of expert-labeled scores. For data-to-text, we test \method on the WebNLG 2017 challenge \cite{gardent2017creating}. For image captioning, we test \method on the COCO image captioning challenge 2015 \cite{Chen2015MicrosoftCC}. We use \citet{freitag2021experts} annotated TED dataset as our development set to select the hyper-parameters in Error Synthesis Models and SEScore Metric Model. We comprehensively analyze each component of our pipeline and their contributions to the final results. 
%\ww{Somehow I think you can say in a footnote that you plan to release the code and synthesize datasets, otherwise such metric paper is not going to be useful.}
% \ww{not sure if you need to justify the chosen language pairs}

\begin{table*}[t]\small
    \centering
    %\begin{tabular}{@{}l|lllllllll@{}}
    \begin{tabular}{@{}l|ccccccccc@{}}
        \toprule
        %& & \multicolumn{4}{c}{\bf English to German} & \multicolumn{4}{c}{\bf Chinese to English} \\
        %\cmidrule(r){3-6}
        %\cmidrule(l){7-10}
        %& & \multicolumn{2}{c}{\bf Newtest2020} & \multicolumn{2}{c}{\bf Newtest2021} & \multicolumn{2}{c}{\bf Newtest2020} & \multicolumn{2}{c}{\bf Newtest2021} \\
        \multicolumn{2}{c}{\multirow{2}{*}{\bf Model Name}} & \multicolumn{2}{c}{\bf WMT20 (En$\rightarrow$De)} & \multicolumn{2}{c}{\bf WMT21 (En$\rightarrow$De)} & \multicolumn{2}{c}{\bf WMT20 (Zh$\rightarrow$En)} & \multicolumn{2}{c}{\bf WMT21 (Zh$\rightarrow$En)} \\
        \cmidrule(r){3-4}
        \cmidrule(l){5-6}
        \cmidrule(l){7-8}
        \cmidrule(l){9-10}
        \multicolumn{2}{c}{} & Kendall & Pearson & Kendall & Pearson & Kendall & Pearson & Kendall & Pearson\\ \midrule
        \parbox[t]{5mm}{\multirow{2}{*}{\rotatebox[origin=c]{90}{\shortstack{With\\HL.}}}}
        & \multicolumn{1}{c}{BLEURT} & 0.229* & 0.476 & 0.052* & 0.383 & 0.218* & 0.531 & 0.078 & 0.423\\
        & \multicolumn{1}{c}{COMET(DA)} & \textbf{0.283} & 0.633 & 0.103 & \textbf{0.650} & 0.256 & 0.628 & \textbf{0.114} & 0.452 \\
        \cmidrule(l){1-10}
        %& \multicolumn{1}{c}{W/o Human Labels} & Kendall & Pearson & Kendall & Pearson & Kendall & Pearson & Kendall & Pearson\\ \midrule
        \parbox[t]{5mm}{\multirow{7}{*}{\rotatebox[origin=c]{90}{\shortstack{W/o\\Human Labels}}}}
        & \multicolumn{1}{c}{TER} & -0.221* & 0.627* & -0.171* & -0.356  & -0.238* & -0.516* & -0.177* & -0.338 \\
        & \multicolumn{1}{c}{BLEU} & 0.112* & 0.322* & 0.010* & 0.358 & 0.120* & 0.562* & 0.030* & 0.330*\\
        & \multicolumn{1}{c}{ChrF} & 0.163* & 0.333* & 0.030* & 0.326 & 0.151* & 0.534* & 0.042* & 0.296*\\
        & \multicolumn{1}{c}{BARTScore} & - & - & - & - & 0.176* & 0.580 & 0.063* & 0.335*\\
        & \multicolumn{1}{c}{BERTScore} & 0.166* & 0.260* & 0.063* & 0.322 & 0.228* & 0.549* & 0.092* & 0.362*\\
        & \multicolumn{1}{c}{PRISM} & 0.208* & 0.219* & 0.068* & 0.198 & 0.240* & 0.505* & 0.101* & 0.352 \\
        & \multicolumn{1}{c}{\method} & 0.273 & \textbf{0.706} & \textbf{0.139} & 0.629 & \textbf{0.261} & \textbf{0.684} & 0.108 & \textbf{0.501}\\
        \bottomrule
    \end{tabular}
    \caption{Segment-level Kendall ($\tau$) and System-level Pearson correlation ($|\rho|$) on En-De and Zh-En for WMT2020 and WMT 2021 Testing sets with Expert-based MQM labels. * indicates that \method significantly outperforms baselines with  p values < 0.05.}%\ww{Supervised vs. Unsupervised?}}
    \label{tab:main-mt}
\end{table*}

\subsection{Pre-training setup}
\label{sec:pretrain_setup}
\paragraph{Synthetic Error Data} We use the  WMT19 \cite{barrault-etal-2019-findings} training News Complimentary dataset \cite{TIEDEMANN12.463} as the raw pretraining data. It contains News articles across 16 different languages.
% In particular, it contains 290K German sentences and 126K English sentences.
We randomly sampled 120K sentences for English and 120K for German, then generated error synthetic sentences from them. 
%We will discuss how quantity of synthetic data impacts the final performance in section~\ref{sec:data_analysis}. 
To adopt to the text domain of WebNLG and Image captioning, we generate 30k and 40k error synthetic sentences from the text portion of the WebNLG \cite{gardent-etal-2017-webnlg} and image captioning's training data \cite{Chen2015MicrosoftCC}. We use those data to train two separate checkpoints for WebNLG and image captioning evaluations. We discuss the effects of cross-domain evaluation in Appendix~\ref{sec:domain_adaptation}.

\paragraph{Error Synthesis Models}
We use four pretrained language models in the error synthesis process. First, we use an mBART model \cite{liu2020multilingual} to generate a span of tokens for the \verb|<mask>| positions for both insertion and replacement operations. Second, we use an XLM-RoBERTa model \cite{conneau2020unsupervised} to predict a token for \verb|<mask>| using MLM's objective for both single token insertion and single token replacement. Finally, we use RoBERTa models fine-tuned on MNLI and XNLI as our entailment classification model for English and German respectively. These two models are used to determine the bidirectional relations of a synthetic sentence and a raw text to measure the severity of the synthetic text. We set the synthesis hyperparameters $\lambda_e=5$, $\lambda_d=1.5$, $\lambda_r=1.5$, and $\lambda_s=4$. We generate all synthesized dataset on one RTX A6000 GPUs. It costs 0.5 hours to generate 10K sentences.

\paragraph{\method Metric Model.}
To ensure the fair comparison and fully demonstrate the power of our pretraining data, \method uses the comparable model size compared to the COMET \cite{rei-etal-2020-comet}. Specifically, we use XLM-RoBERTa Large as the backbone for our German metric model and RoBERTa Large for English metric model. We use Adam optimizer \cite{kingma2017adam} and set batch size, learning rate and dropout rate of 8, 3e-5 and 0.15 respectively. We use mean squared error to train the metric model. We select the best checkpoint based on the highest Kendall correlation on the TED validation. We include detailed training process and hyperparamters in the Appendix \ref{sec:appendix_sescore_model}. 

\subsection{Baseline Methods}
For machine translation evaluation, we include three WMT baseline methods and five best performed learned metrics. They are (1) Ngram- and distance-based metrics (BLEU \cite{papineni2002bleu}, ChrF \cite{popovic-2015-chrf} and TER \cite{snover-etal-2006-study}); (2) learned metrics requiring human rating data (COMET \cite{rei-etal-2020-comet}, BLEURT \cite{sellam2020bleurt}); (3) learned metrics without human rating data (PRISM \cite{thompson-post-2020-automatic}, BARTScore \cite{Yuan2021BARTScoreEG} and BERTScore \cite{bert-score}). For WebNLG evaluation, we include the three baselines in prior work \cite{gardent-etal-2017-webnlg}: METEOR \cite{banerjee-lavie-2005-meteor}, TER, BLEU, and two learned metrics MoverScore \cite{zhao2019moverscore} and BERTScore. For image captioning, we include five baseline models in the COCO image captioning challenge 2015 \cite{Chen2015MicrosoftCC}: BLEU, METEOR, ROGUE-L \cite{lin-2004-rouge}, CIDEr \cite{vedantam2015cider} and CHrf. We further include  BARTScore and BERTScore and one top-performing task-specific learned metric, LEIC \cite{cui2018learning}.
For all the learned metrics with variants, we choose their checkpoints based on their paper recommendations. We discuss the details of the baseline model setups in the Appendix \ref{sec:baselines}.

\subsection{Evaluation Procedure}

\paragraph{Machine Translation Task}

As WMT20's standard practice \cite{mathur-etal-2020-results}, we compute the correlations of each evaluation metric to the segment- and system- level human scores, on WMT20 and WMT21, with MQM-based labels \cite{freitag2021experts}. For the segment-level correlation, we adopt the Kendall $\tau$ correlation from WMT20 to evaluate the relative rankings between segments of the different systems. For the correlation of system-level scores, we average \method for all reference-candidate pairs of each machine translation system and estimate the absolute Pearson correlation $|\rho|$ to the system-level human judgement scores. 

\paragraph{Data-to-Text Task}
Following the WebNLG challenge \cite{gardent-etal-2017-webnlg}, we use Kendall correlation to evaluate the segment-level correlation. Each generated output is annotated by three aspects: semantics, grammar and fluency. Since our \method is the overall score of accuracy and fluency, we average three aspects of human ratings into one overall score and evaluate segment-level Kendall correlation of the \method to the overall human judgement score.

\paragraph{Image Captioning Task} Following \citet{bert-score}, we compute \method for all reference-candidate pairs of each image captioning system and average all the scores for each system to generate the system-level scores. We compute the system-level Pearson correlation with M3 system-level human judgement score in COCO image captioning challenge \cite{Chen2015MicrosoftCC}. M3 human judgement measures the average correctness of the captions on a scale 1-5.  
The detailed task, data information and evaluation procedures are included in the Appendix \ref{sec:eval_procedure}. 

\subsection{Results on Machine Translation}
In \cref{tab:main-mt}, we show our evaluation results on En-De and Zh-En in both WMT20 and WMT21. 

\paragraph{English to German} We first contrast \method with three WMT baselines (BLEU, TER and Chrf). \method outperforms them significantly in both system-level Pearson and segment-level Kendall correlations. \method shows its superior performance over two recent unsupervised learned metrics (Bertscore and PRISM) leading by an average 8\% and 7\% segment-level Kendall correlation in two years' testing sets. Compared to the supervised models, \method has around 4.4\% improvement in the Kendall correlations at WMT20 and 8.8\% at WMT21 against BLEURT.  Most importantly, \method outperforms the SOTA supervised metric, COMET, by 3.6\% in Kendall for WMT21 and 7.3\% in system-level Pearson correlation. 

\paragraph{Chinese to English} Similar to En-De, \method outperforms three WMT baseline models (BLEU, TER and Chrf) by the great margin in both system-level and segment-level correlations of two years' testing sets. Compared to three strong unsupervised learned metrics, BERTScore, BARTScore and PRISM, \method can outperform them by 4.6\% on average in Kendall correlation in WMT20 and average 2.3\% in WMT21. Compared to the supervised models, we have 4.3\% improvement in the Kendall correlations at WMT20 and 3\% at 
WMT21 against BLEURT. This is significant as BLEURT is previously trained as an English-oriented metric with millions of synthetic data and 5 year's human rating data (WMT15-19). Moreover, \method outperfoms the SOTA supervised COMET model for both segment-level and system-level correlation in WMT20. The remaining gaps of Kendall correlations to the COMET is within 1\%. 

\paragraph{Takeaways:} Machine translation results in En-De and Zh-En demonstrate \method's superior performance to unsupervised metrics and competitive performance against supervised SOTA metrics. 

\subsection{Results on WebNLG Challenge}
\label{sec:webnlg}

% \begin{figure}[t]
%     \includegraphics[width=0.8\textwidth, ]{figures/data_size_feb5.pdf}
%     \caption{Effects of Data Quantity}
%     \label{fig:data}
% \end{figure}

% \begin{figure}[t]
%     \includegraphics[width=0.8\textwidth, ]{figures/funct_types.pdf}
%     \caption{Effects of the Error Types: We use N - 1 function types for the data synthesis process. M represents MBart and X represents XLM-RoBERTa}
%     \label{fig:funct_type}
% \end{figure}

%\paragraph{Performance Analysis} 
\cref{tab:webnlg} shows our segment-level Kendall correlation results for WebNLG Challenge.  \method can outperform three baseline models (Meteor, TER and BLEU) significantly. When comparing to the learned metrics, \method outperforms BARTScore and MoverScore significantly by leading 8.2\% and 3\% improvements on Kendall correlations. Moreover, it improves the top-performing unsupervised metric, BERTScore, by 0.3\%.

\subsection{Results on Image Captioning Challenge}
\label{sec:captioning}
\cref{tab:caption} demonstrates our system-level Pearson correlation results for the COCO image captioning challenge. \method outperforms all task-agnostic and task-specific baseline metrics. The correctness metric in image captioning creates a challenge evaluation scenario, such that evaluating only on semantic coverage does not cover all model mistakes. Metrics including METEOR, BLEU, even BERTScore with pretrained word embeddings only yield weak or moderate correlations to the human judgements. \method further outperforms significantly to BERTScore with idf weights and BARTScore which covers faithfulness. Most importantly, \method outperforms two task-specific metcis, LEIC \cite{cui2018learning} and CIDER \cite{vedantam2015cider}. by 6.1\% and 1.8\% Pearson correlations. This is a significant result, as LEIC is a trained metric that takes image as additional inputs, optimized on the COCO data distributions and CIDER is a consensus based evaluation purely used for image descriptions.

\textbf{Takeaways:} Results in \cref{sec:webnlg} and \cref{sec:captioning} verify our prior assumptions that  despite our synthesized error types are originated for Machine Translation tasks, they are useful and applicable to multiple domains and tasks. As benefited from the reference-only evaluation setup, our pretrained evaluation metric can correlate well to the human judgements in various text generation settings, e.g with or without requiring source data to be text.

\begin{table}
\resizebox{\textwidth}{!}{%  
\begin{floatrow}
\capbtabbox{
        \resizebox{.42\textwidth}{!}{%  
        \begin{tabular}{@{}ll@{}}
            \toprule
            \multicolumn{2}{c}{\bf WebNLG} \\
            \cmidrule(r){1-2}
            \multicolumn{1}{c}{Model Name} & Kendall \\ \midrule
            \multicolumn{1}{c}{METEOR} & -0.388*\\
            \multicolumn{1}{c}{TER} & -0.345*\\
            \multicolumn{1}{c}{BLEU} & 0.289*\\
            \multicolumn{1}{c}{BARTScore} & 0.317* \\
            \multicolumn{1}{c}{MoverScore} & 0.369*\\
            \multicolumn{1}{c}{BERTScore} & 0.396\\
            \midrule
            \multicolumn{1}{c}{\method} & \textbf{0.399}\\
            \bottomrule
        \end{tabular}
        }
}{
 \caption{Segment-level Kendall Correlation ($\tau$) on WebNLG 2017. * indicates that \method significantly outperforms baselines with p value < 0.05.}
 \label{tab:webnlg}
}
\capbtabbox{
        \resizebox{.42\textwidth}{!}{%  
        \begin{tabular}{@{}ll@{}}
            \toprule
            \multicolumn{2}{c}{\bf COCO Image Captioning} \\
            \cmidrule(r){1-2}
            \multicolumn{1}{c}{Model Name} & Pearson \\ \midrule
            \multicolumn{1}{c}{METEOR} & 0.349* \\
            \multicolumn{1}{c}{CHrF} & 0.442\\
            \multicolumn{1}{c}{BERTScore} & 0.459*\\
            \multicolumn{1}{c}{ROGUE-L} & 0.589*\\
            \multicolumn{1}{c}{BLEU} & 0.605\\
            \multicolumn{1}{c}{BERTScore(Idf)} & 0.644\\
            \multicolumn{1}{c}{BARTScore} & 0.688\\
            \multicolumn{1}{c}{LEIC+} & 0.720\\
            \multicolumn{1}{c}{CIDER+} & 0.763\\
            \midrule
            \multicolumn{1}{c}{\method} & \textbf{0.781}\\
            \bottomrule
        \end{tabular}
        }
}{
 \caption{\footnotesize System-level Pearson Correlation ($|\rho|$) on COCO Image captioning's M3 Metric. Metrics with + are directly cited from \citet{cui2018learning}. * indicates that \method significantly outperforms baseline models with p value < 0.05}
 \label{tab:caption}
}
\end{floatrow}
}
\end{table}

\section{Quantitative Analysis}
To validate the proposed \method training technique, we analyze the effects of  data quantity, the stratified components, and synthetic error types. We include the cross-domain evaluation in the Appendix \ref{sec:domain_adaptation}. We include a detailed qualitative analysis of \method regarding to its robustness and limitations in Appendix \ref{sec:qualitative}.

\begin{figure*}[t]
\begin{floatrow}\small
     \centering
     \capbtabbox[\textwidth]
    {
        \begin{tabular}{@{}lllllllll@{}}
            \toprule
            \multirow{1}{*} &  \multicolumn{2}{c}{\bf WMT20 (En$\rightarrow$De)} & \multicolumn{2}{c}{\bf WMT21 (En$\rightarrow$De)} & \multicolumn{2}{c}{\bf WMT20 (Zh$\rightarrow$En)} & \multicolumn{2}{c}{\bf WMT21 (Zh$\rightarrow$En)} \\
            \cmidrule(r){2-3}
            \cmidrule(l){4-5}
            \cmidrule(l){6-7}
            \cmidrule(l){8-9}
            \multicolumn{1}{c}{Stratified Components} & Kendall & Pearson & Kendall & Pearson & Kendall & Pearson & Kendall & Pearson\\ \midrule
            \multicolumn{1}{c}{\method w/o synthesized error} & 0.129 & 0.204 & 0.004 & 0.457 & 0.180 & 0.569 & 0.044 & 0.364\\
            \multicolumn{1}{c}{\method w/o severity measures} & 0.249 & 0.549 & 0.103 & 0.608* & 0.234 & -0.058 & 0.097 & 0.278\\
            \multicolumn{1}{c}{\method} & \textbf{0.273} & \textbf{0.706}* & \textbf{0.139} & \textbf{0.629}* & \textbf{0.261} & \textbf{0.684}* & \textbf{0.108} & \textbf{0.501}\\
            \bottomrule
        \end{tabular}
    }
    {
        \caption{Abalation study on the stratified error synthesis on En-De and Zh-En for WMT2020 and WMT 2021 Testing sets with Expert-based MQM labels. * indicates the Pearson correlation has  p values < 0.05.}
        \label{tab:severe_abalation}
    }
    %\hspace{.005\textwidth}
\end{floatrow}
\end{figure*}

\begin{figure*}[t!]
    \centering
    \begin{subfigure}{.24\textwidth}
    \includegraphics[width=1\textwidth]{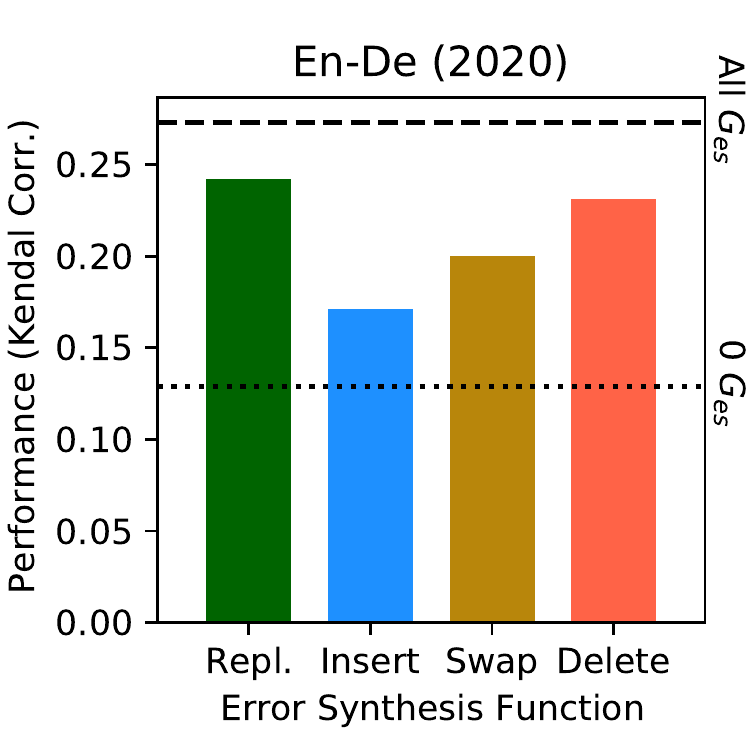}
    \end{subfigure}
    \begin{subfigure}{.24\textwidth}
    \includegraphics[width=1\textwidth]{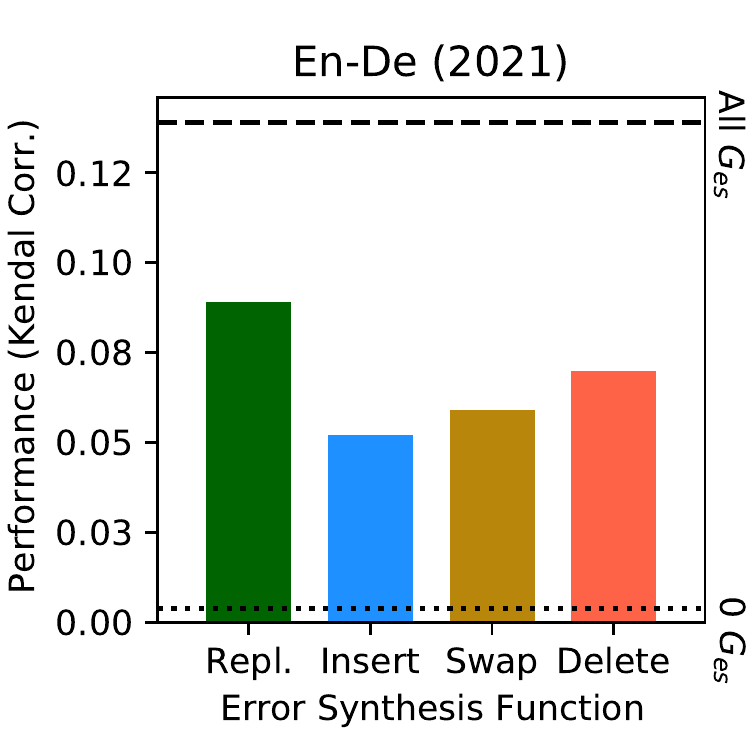}
    \end{subfigure}
    \begin{subfigure}{.24\textwidth}
    \includegraphics[width=1\textwidth]{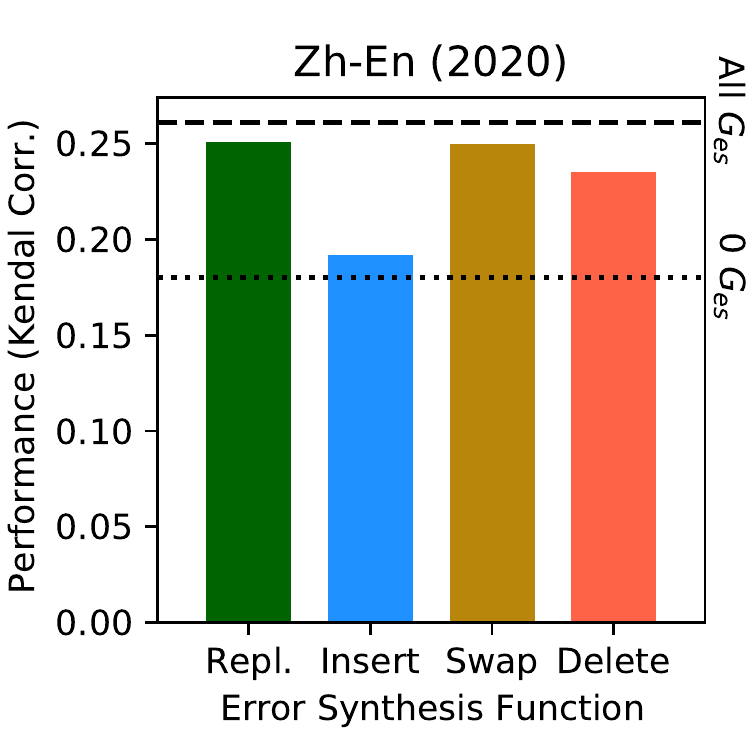}
    \end{subfigure}
    \begin{subfigure}{.24\textwidth}
    \includegraphics[width=1\textwidth]{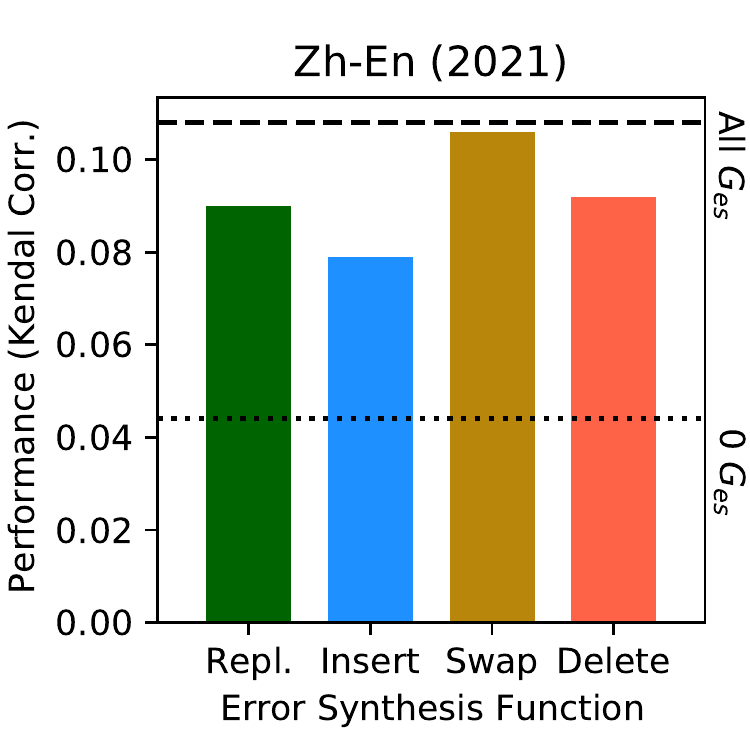}
    \end{subfigure}
    \caption{Effects of the error types: demonstrating the results achieved when \textbf{\textcolor{repgreen}{Replace}}, \textbf{\textcolor{insblue}{Insert}}, \textbf{\textcolor{swapyell}{Swap}}, or \textbf{\textcolor{delred}{Delete}} is separately applied. Dashed line (All $g_{es}$) represents the aggregate performance when all four synthesis functions are used together. The dotted line (0 $g_{es}$) represents the baseline performance of \method when none of the error synthesis functions are applied.}
    \label{fig:funct_type}
\end{figure*}

% \subsection{Data Quantity Effects}
% \label{sec:data_analysis}
% % Previous studies \cite{sellam2020bleurt} found that synthetic dataset in orders of magnitudes of human annotated data is important to learn a useful metric. However, determining the right amount is crucial for the resources consuming process.
% We use 10k, 20k, 40k and 120k synthetic error samples to train \method models and evaluate their Kendall correlations on WMT20. We observe that the Kendall correlation reaches an optimal level at 40k synthetic sentences in Zh-En and 10k synthetic sentences in En-De. This demonstrates the potential gap between synthetic and real error distributions. It also indicates that the optimal performance can be achieved through error perturbations with small amount of raw text.

\subsection{Effects of the Stratified Components}
%There are two components in our stratified process: 1) Each synthetic error sentence is generated through incremental error injections 2) Each error is estimated separately on its severity level. 
To study the effects of each component, we include the \method w/o synthesized error \footnote{We mean-pooled the word embeddings from pretrained models \cite{conneau2020unsupervised, Liu2019RoBERTaAR} to generate each sentence embedding and compute the cosine similarities of the sentence embeddings for evaluation.} and \method with without severity measures \footnote{we remove the severity scoring component in \method by assigning all errors to be minor, with score -1. The final score will be within 0 to -5. We use this new score labeling to pretrain a \method without severity measures.}. In \cref{tab:severe_abalation}, we demonstrate that \method 
without severity measures can still achieve the strong performance improvements over the base language model, leading average 11\% and 5\% in segment-level Kendall correlation at En-De and Zh-En, respectively. This result demonstrates that our incremental injection of synthetic errors can achieve high human correlations on the segment-level rankings, providing the first layer of our stratified process. However, without severity measures, \method can hardly determine system level ranking, indicating by weak system-level correlations in Zh-En. By adding the severity measures into our stratified pipeline, we observe a large system-level correlation improvements in both En-De and Zh-En. The segment-level Kendall correlation can be further improved by average 3\% in En-De and 2\% in Zh-En. This study demonstrates the effectiveness and importance of our stratified components in both segment-level and system-level correlations to human judgements.

\subsection{Effects of the Error Types}
To understand each error type's contribution to the final pretraining outcomes, we use each error synthesis function to generate separate synthesized data and use each data to train a \method. We benchmark \method's performance with each error synthesis function in both years' language directions. \cref{fig:funct_type} demonstrates that individual error synthesis function contributes to the pretrained metric differently in different language directions. 

In \cref{fig:funct_type}, from both En-De and Zh-En, we observe that all four error synthesis functions are effective as they bring up the base Kendall performance of at least 5\% from En-De and at least 7\% from Zh-En in both year's testing sets. We observe that the Replace and Delete tasks are the two prominent error synthesis functions in both En-De and Zh-En. 
On the contrary, the insert operation has the relatively minor effects in both En-De and Zh-En. Our best assumption is that large pretrained language model tends to produce semantically close content when giving the full context of the sentence. Therefore, most of insert produced errors are relatively minor and are not able to simulate Addition error types under diverse severity levels. Lastly, we observe that the swap operation has different effects in different language directions. From Zh-En, the \method trained solely on Swap errors can achieve equal to or less than 1\% Kendall correlations compared to the \method with four different operations. However, in En-De, the swap function only has moderate effects.

\textbf{Takeaways:} We demonstrate that all error synthetic functions can improve Kendall correlations to the human judgements. However, the effect of each error synthetic functions is related to the actual error distributions in each task. Aggregating all four error synthetic functions should be considered to achieve a general error distributions which is robust to different domains or tasks.

\section{Conclusion}
To conclude, we introduced SEScore, a reference-based metric for text generation evaluations. Without human labels, SEScore can outperform all unsupervised evaluation metrics and achieve competitive performance to the SOTA supervised approaches. We demonstrate that our stratified error synthesis approach makes model aware of individual errors with different severity levels, achieving high correlation to the human judgements.

\section{Acknowledgements}
This work was supported by the National Science Foundation award \#2048122. The views expressed are those of the author and do not reflect the official policy or position of the US government. We thank the Robert N.Noyce Trust for their generous gift to the University of California via the Noyce initiative.

\section{Ethics and Limitations}
Our qualitative analysis in Appendix \ref{sec:qualitative} highlights three main limitations in the \method framework. First, we observe that it is difficult for \method to detect punctuation errors. 
%For example, a generated sentence ``time to eat grandma'' means something very different from ``time to eat, grandma'' but \method is not sensitive to this nuance, 
As they are not represented in the entailment data distributions. 
Second, \method disagrees with human judgements when human annotations contain uncertainties (e.g., high inter-rater disagreement on the severity of an error). 
Perhaps in these cases human opinions are too inherently subjective to model well in the first place. 
Regardless, \method is not likely to produce rankings exactly matching human annotators when human rating difference is less than 1. 
Lastly, \method disagrees more heavily with human annotators on the quality of long generated text passages. 
We assumed that this is due to our limited sentence embedding space while individual errors will be mitigated by the long sentence contexts. Most importantly, we observed that those three limitations are also commonly occurred in the three top-performing baseline metrics (BERTScore \cite{bert-score}, PRISM \cite{thompson-post-2020-automatic} and COMET \cite{rei-etal-2020-comet}), motivating more future works to investigate on those issues. We demonstrate \method's superior performance over other baselines. However, \method can not be used to replace human judgements. We will support two frameworks of SEScore: SEScore with and without severity measures. SEScore with severity measures can support up to 100 languages that XLM-Roberta pretrained on. SEScore with severity measures currently supports 15 languages: English, French, Spanish, German, Greek, Bulgarian, Russian, Turkish, Arabic, Vietnamese, Thai, Chinese, Hindi, Swahili and Urdu. All code and synthesized data samples will be publicly released following deanonymization. 
% Also move E.2 Limitations from Appendix here

\bibliography{anthology,custom,main}
\bibliographystyle{acl_natbib}

\newpage
\appendix
\label{sec:appendix}
\newpage 
 \section{Algorithm Details}
\begin{algorithm}[h!]\small
  \caption{Stratified Error Synthesis}
  \label{alg:SSE}
  \KwIn{Seed sentence set $S = \{x_1, x_2, ... , x_n\}$, $\lambda_e$, $\lambda_d$, $\lambda_r$, $\lambda_s$, editing model set $M$.}
  \KwOut{Synthetic reference and error text $D$.}
%   $\sum_{1}^{n} (\{(x_i, y'_i ), s'_i\})$, where $y_i$ is synthesized error sentence and $s'_i$ is the pseudo rating
  $D=\emptyset$\;
  \For{$i = 1 .. n$}{
    $l = len(x_i)$, 
    $y_{new} = x_i$, $s_i=0$\;
    $k \sim Poisson(\lambda_e)$\; 
    \For{$j = 1 .. k$}{
        $y_{\text{old}}=y_{\text{new}}$\;
        $\mathrm{edit} \sim \mathrm{Random}(\{\mathrm{Ins}, \mathrm{Del}, \mathrm{Rep}, \mathrm{Swap}\})$\;
        \Switch{$\mathrm{edit}$}{
            \Case{\textup{Ins}}{
                sampling $h \sim \mathrm{Uniform}(0, l)$ s.t. $h$ does not overlap the previous edited spans\;
                Randomly select a model from $M$ to generate a phrase $f$ to insert at position $h$ of $y_{\text{new}}$\;
            }
            \Case{\textup{Del}}{
                \Repeat{the span from $h$ to $b+ll-1$ does not overlap the previous edited spans}{
                    draw $h \sim \mathrm{Uniform}(0,l)$\;
                    draw $ll \sim \mathrm{Poisson}(\lambda_d)$\;
                }
                Remove a span of length $ll$ at position $h$ from  $y_{\text{new}}$\;
            }
            \Case{\textup{Rep}}{
                \Repeat{the span from $h$ to $b+ll-1$ does not overlap the previous edited spans}{
                    draw $h \sim \mathrm{Uniform}(0,l)$\;
                    draw $ll \sim \mathrm{Poisson}(\lambda_r)$\;
                }
                Randomly select a model from the model base $M$ to generate a phrase $f$\;
                Replace the segment of $y_{\text{new}}$ from $h$ to $h+ll-1$ with $f$\;
            }
            \Case{\textup{Swap}}{
                \Repeat{the span from $b$ to $b+ll-1$ does not overlap the previous edited spans}{
                    draw $h \sim \mathrm{Uniform}(0,l)$\;
                    draw $ll \sim \mathrm{Uniform}(1..\lambda_s)$\;
                }
                Swap the tokens in $y_{\text{new}}$ at positions $h$ and $h+ll$\;
            }
        }
        $s_i += \text{$S_{es}$}(y_{old}, y_{new})$\;
    }
    $D \leftarrow D \cup \{(x_i, y_{new}, s_i) \}$\;
  }
%     Initialize an array $q_i$ with length $k$ and values of $k-1, k-2, ..., 0$ \\
%     Draw a Possion distribution $\lambda=1.5$ from 1 to 5 for the number of error iterations on $x_i$ \\
%     \For{each iteration of text $x_i$}
%     {
%         Randomly select an error function, $f_{es}$ from $insert$, $replace$, $delete$ and $swap$ \\
%         Randomly select a start index from tokenized length, $j$ \\
%         \If{$f_{es} == insert$ and $q_i[j] > 0$} {
%             Insert a $<mask>$ token after j \\
%             Use MBart or Roberta to predict the $<mask>$ \\ 
%             update all tokens on the left of j with $min(j-1, k-1)$
%         } 
%     }
%   }
 \end{algorithm}
 
 \section{Implementation Details of the Pretraining Pipeline}
 
 This section provides the implementation details for both error synthesis models and SEScore metric model.  
 
% \subsection{Error Synthesis Models}
% \label{sec:appendix_error_syn_model}

\subsection{SEScore Metric Model}
\label{sec:appendix_sescore_model}
The feed-forward hidden dimensions are 2048 and 1024. We use \verb|tanh| as our activation function. The training process takes 1, 3, 2 and 1 epoches for machine translation Zh-En,  machine translation En-De, WebNLG and image captioning, respectively.

\section{Experiments-Supplementary Material}

\subsection{Details about the Baseline Models}
\label{sec:baselines}
For all model variants, we choose each model based on two criteria: their paper recommendations and comparable model size to SEScore. 

For BERTScore \cite{bert-score}, we follow its model recommendation by using roberta-large for English texts and bert-base-multilingual-cased for German texts. For all BERTScore in the paper, we report their F1 scores. For BLEURT \cite{sellam2020bleurt}, we use BLEURT-Large (Max token 128, 24 layers and 1024 hidden units, comparable size to SEScore) for English texts and BLEURT-20-D12 for German texts. For COMET \cite{rei-etal-2020-comet}, we choose their best checkpoint wmt20-comet-da (exactly the same model size to SEScore) to evaluate its performance. We use bart-large-cnn to evaluate BARTScore \cite{Yuan2021BARTScoreEG}'s performance. We NLTK \cite{bird2009natural} library to implement BLEU \cite{papineni2002bleu}, METEOR \cite{banerjee-lavie-2005-meteor}, CHrF \cite{popovic-2015-chrf} and ROUGE-L \cite{lin-2004-rouge}. We report LEIC \cite{cui2018learning} and CIDEr \cite{vedantam2015cider}'s performance through prior study \cite{cui2018learning}.    

\subsection{Details about the Evaluation Procedures and Test Data Information}
\label{sec:eval_procedure}

\paragraph{Machine Translation Task} We use WMT20 and WMT21 \cite{50871} 's testing sets (Newtest2020 and Newtest2021), with mqm-based expert labels, as our main evaluation corpus. WMT20 (Chinese$\rightarrow$ English) contains 2000 segments across 155 documents and WMT (English$\rightarrow$ German) contains 1418 segments across 130 documents, respectively. WMT21 (Chinese$\rightarrow$ English) contains 1948 segments and WMT21 (English$\rightarrow$ German) contains 1002 segments, respectively. There are two types of human judgement scores: Segment-level and System-level scores. Segment-level human judgement score assigns a single score to each reference-candidate pair. System-level human judgement score assigns a single score to each system based on all \{reference, system output\} pairs. We follow the WMT20's standard practice to evaluate metric performance using both system-level and segment-level correlation.

For \textbf{system-level evaluation}, we average SEScore for all reference-candidate pairs of each machine translation system and estimate the absolute Pearson correlation $|\rho|$ to the System-level human judgement scores. \citet{50871} annotated top 10, 10, 17 and 15 top performing systems of En-De and Zh-En in Newtest2020 and En-De and Zh-En in Newtest2021, respectively.

% In the original WMT 2020 shared metric \cite{mathur-etal-2020-results}, they discard pairwise rankings when annotations are missing or raw scores are differed by less than 25. However, in MQM expert-based labels, since data is high-precision with fine-grained error annotations, small difference between two segments is significant for the precise relative ranking.

For \textbf{segment-level evaluation}, we adopt the Kendall $\tau$ correlation from WMT20 \cite{mathur-etal-2020-results} to evaluate the relative rankings between segments of the different systems (See Eqn \ref{eq:kendall}). Following the prior study's suggestion \cite{freitag2021experts}, we use the absolute threshold between two segment scores to determine the relative rankings of both En-De and Zh-En.
To prepare all the relative ranking pairs for Kendall correlation, we removed all the pairs which have the exactly same annotations and cleaned erroneous texts. In the end, we have 76,087 pairs from Zh-En and 54405 pairs from En-De in Newtest2020 and 38758 pairs from En-De and 52498 pairs from Zh-En in Newtest2021. 

The \textbf{Kendall's Tau-like formulation} is defined as following:

\begin{equation}
  \tau = \frac{Concordant - Discordant}{Concordant + Discordant}
  \label{eq:kendall}
\end{equation}

where Concordant indicates the number of the correct predictions in the pairwise ranking and Discordant indicates the number of the misrankings. 

\paragraph{Data-to-Text Task} The WebNLG dataset \cite{gardent-etal-2017-webnlg} consists a set of data extracted from DBpedia and requires systems to map entities (e.g., buildings, cities, artiests) to text. We use 9 submissions for WebNLG challenge. Each system generates 223 outputs. In total, we have 4,677 output sentences. Following the WebNLG challenge \cite{gardent-etal-2017-webnlg}, we use Kendall $\tau$ correlation to evaluate the relative rankings between segments of the different systems. From combinations of rankings and data cleaning, we obtain 7725 relative ranking pairs. Each generated output is evaluated by three aspects: semantics, grammar and fluency. Since our SEScore is the overall score of accuracy and fluency, we average three aspects of human ratings into one overall score and evaluate segment-level Kendall correlation of the SEScore to the overall human judgement score. The Kendall $\tau$'s formulation is shown in Eqn \ref{eq:kendall}.

\paragraph{Image Captioning Task} COCO 2015 Captioning Challenge \cite{Chen2015MicrosoftCC} consists of the human judgements from the 11 submission entries \footnote{There are 15 submission entries in the COCO 2015 Captioning Challenge \cite{Chen2015MicrosoftCC}. However, 3 entries did not submit their validation outputs and 2 systems have the identical validation outputs. Therefore, we use the submissions from the 11 entries}. Following the prior study \cite{cui2018learning, bert-score}, we perform our experiments on the COCO validation set, as we do not have access to COCO test set where human judgements were performed. Using the findings of the prior works \cite{cui2018learning, bert-score}, we argue that the human judgements on the validation set are sufficiently close to the ones on the testing set.  

\section{Quantitative Analysis}

\subsection{Effects of the Cross Domain Evaluation}
\label{sec:domain_adaptation}
As domain shifts have been repeatedly reported by the previous studies \cite{sellam2020bleurt, Yuan2021BARTScoreEG}, we conduct experiments to study SEScore before and after domain adaptation in WebNLG and image captioning. In Table \ref{tab:abalation_domain}, due to the close data distribution and error types in WebNLG and machine translation, we find that SEScore pretrained on machine translation error synthetic data can achieve strong cross-domain performance in WebNLG and competitive to in-domain pretrained variant. However, when larger domain difference presents between machine translation and image captioning, domain adaptation plays a major role by leading metric from no correlation of cross-domain performance to high correlation to human judgements. This finding suggests that our domain adaptation strategy is effective in adapting synthetic error sentences into different domains cross several NLG tasks. This technique can provide major benefits in training a powerful learned metrics in narrowed domain, e.g low resource language of machine translation.   

\begin{table}
\resizebox{\textwidth}{!}{%
\begin{floatrow}\small
     \centering
     \capbtabbox[\textwidth]
    {
        \begin{tabular}{@{}lll@{}}
            \toprule
            \multicolumn{1}{c}{Task} & WebNLG ($\tau$) & COCO ($\rho$) \\ \midrule
            \multicolumn{1}{c}{Cross-domain Performance} & 0.396 & -0.0428\\
            \multicolumn{1}{c}{In-domain Pretraining} & \textbf{0.399} & \textbf{0.781}\\
            \bottomrule
        \end{tabular}
    }
    {
        \caption{Abalation study on the cross-domain evaluation at WebNLG and COCO image captioning Challenge.}
        \label{tab:abalation_domain}
    }
    %\hspace{.005\textwidth}
\end{floatrow}
}
\end{table}

% \subsection{Effects of the Data Quantity}
% \begin{figure}[h!]
%     \centering
%     \includegraphics[width=.95\linewidth]{figures/1_data_quant.pdf}
%     \caption{Relationship between data quantity and performance ($\tau$) for Zh-En and En-De translation.}
%     \label{fig:data}
% \end{figure}

\section{Qualitative Analysis} 
\label{sec:qualitative}
We study the outputs of three best performing baseline models (BERTScore, PRISM, COMET) and SEScore on WMT20 Chinese-to-English. Ideally, the rankings produced by the automatic evaluation metrics should be similar to the rankings assigned by the human score.

\subsection{Robustness Analysis} \hyperref[tab:goodcasestudy]{Table 7} shows examples where SEScore disagree largely to the baseline models (BERTScore and PRISM) about the pairwise rankings. We observe that SEScore is effective on distinguishing pairs, which are differed on only one minor error, demonstrated at case No.1 in \hyperref[tab:goodcasestudy]{Table 7}, BERTScore is extremely vulnerable in such cases, since BERTScore's approach relies largely on the overall semantic coverages of the word embeddings. Minor mistake, like inappropriate use of "subscribers" is hard to reflect to in its overall score. We observe the similar shortcomings in PRISM and COMET. We investigate the robustness of the word order for all automatic evaluation metrics (Case No.2). Similar to the previous findings \cite{Sai_2021_EMNLP}, BERTScore suffers greatly when word order is shuffled and fails to capture the shifts in semantic meanings. All PRISM, COMET and SEScore are able to give the correct rankings. Case No.3 and No.4 demonstrate the metrics' capabilities in distinguishing the severe and minor errors. For example, in "Worse" sentence of case No.3, although "Chinese citizens are becoming more and more convenient to apply for visas " shares a lot word coverage to the reference, it completely alters the sentence meaning. According to the MQM-based human evaluation criteria \cite{freitag2021experts}, this is a severe error and should be labeled as -5. However, due to their evaluation criteria, both PRISM and BERTScore are incapable in distinguishing such differences. In this analysis, we demonstrate qualitatively that SEScore's superior performance over unsupervised top-performing metrics (BERTScore and PRISM) and comparative performance to the SOTA supervised metric COMET. Moreover, SEScore demonstrates its better score alignments to the human judgements against other metrics. Its scores are directly interpretable under MQM expert-based human evaluation framework \cite{freitag2021experts}.

\subsection{Limitations} 
\hyperref[tab:badcasestudy]{Table 8} shows examples where SEScore disagrees with human judgements about the pairwise rankings. We observe that SEScore find it difficult to detect punctuation errors. For example, SEScore fails to correctly rank No.1 where "Worse" example's punctuation has higher severity error. Second, SEScore disagrees with human judgements when human labels contains uncertainties (Human annotators do not have the agreements on the severity measures), indicating by No.2 and No.3. With the close severity differences (<1 human rating difference), SEScore is not likely to produce rankings exactly matching human annotators. Lastly, for the long text generation with more than 100 words (No.4), we observe that SEScore fails to produce correct rankings or align to the human judgements. We assumed that this is due to our limited sentence embedding space while individual errors will be mitigated by its long sentence contexts. Moreover, we observed that those three limitations are also commonly occurred in the three top-performing baseline metrics (BERTScore, PRISM and COMET), motivating more future works to investigate on those issues.   

% \newpage

% \quad

\begin{table*}
\centering\small
\begin{adjustbox}{angle=270}

        \begin{tabular}{lp{7.25cm}p{7.25cm}llllll}
            \toprule
            & & & & BERT- & & & SEScore & \\
            \multicolumn{1}{c}{No.} & Reference  & Model Outputs & Category & Score & PRISM & COMET & (\textit{Ours}) & Human\\ \midrule
            \multicolumn{1}{c}{1} & However, Boston Dynamics pointed out that Spot has now entered mass production, and most of its subscribers were construction and energy companies. & However, \underline{Boston Dynamics Technology} pointed out that Spot has entered the stage of mass production, and most of the buyers are construction and energy operators. & Better & 0.960 & -1.410 & 0.282 & \textbf{-4.435} & -4.333\\\\
            
            \multicolumn{1}{c}{1} &  & However, \underline{Boston Power Technology} pointed out that Spot has now entered the stage of mass production, and most of the \underline{subscribers} are construction and energy companies. & Worse & 0.964 & -0.975 & 0.695 & \textbf{-5.934} & -6.000\\\\\midrule
            
             \multicolumn{1}{c}{2} & said the person mentioned above. &  \underline{The above-mentioned person} said. & Better & 0.901  & \textbf{-1.345} & \textbf{0.308} & \textbf{-2.112} & -2.000\\\\
             
             \multicolumn{1}{c}{2} &  & The person \underline{mentionedThe above} said. & Worse & 0.922  & \textbf{-2.921} & \textbf{-0.650} & \textbf{-5.287} & -5.333\\\\\midrule
             
             \multicolumn{1}{c}{3} & In addition to visa-free and visa-on-arrival arrangements, it is becoming more convenient for Chinese citizens to apply for visas, and the procedures are becoming simpler. & In addition to visa exemption, landing \underline{visavisa} and other arrangements, it is more and more convenient for Chinese citizens to apply for visas and the procedures are more and more simplified. & Better & 0.939  & -1.213 &  \textbf{0.808} & \textbf{-1.827} & -1.733\\\\
             
             \multicolumn{1}{c}{3} &  & In addition to visa-free and visa-on-arrival arrangements, \underline{Chinese citizens are becoming more and more convenient} \underline{to apply for visas}, and their procedures are becoming more and more simplified. & Worse & 0.976  & -0.675 & \textbf{0.576} & \textbf{-6.301} & -6.667\\\\\midrule

             \multicolumn{1}{c}{4} & The mobile phone client highlights the artificial intelligence voice function, adapts to the trend of mobile transmission, and provides users with the carry-on “the Story of China” players. & The mobile app features AI speech \underline{recognition} in accordance with the trend of mobile communication and provides users with a portable "\underline{China Story}" player. & Better & 0.921  & -2.590 & \textbf{0.910} & \textbf{-0.046} & -1.700 \\\\
             
             \multicolumn{1}{c}{4} & & \underline{Mobile phone clients} highlight the voice function of artificial intelligence, adapt to the trend of mobile communication, and provide users with portable "\underline{China Good Story}" players. & Worse & 0.938  & -1.513 & \textbf{0.646} & \textbf{-5.004} & -5.700\\\\
            \bottomrule \\
            \multicolumn{9}{p{24.5cm}}{Table 7: Example sentences in 4 relative ranking pairs assigned by BERTScore(F1), PRISM, COMET, SEScore and Human. We use ``Better" and ``Wose" to indicate the model outputs with higher and lower human ratings, respectively. We include all metric outputs and human labels on the right side of the Table. Our SEScore shows its strong correlation to the human judgements while unsupervised baseline models disagree with human ratings significantly. We \textbf{bold} the metric results which produce the correct pairwise rankings. We also \underline{underline} the error spans in each model output.} 
        \end{tabular}
        %\caption{}
        \label{tab:goodcasestudy}

\end{adjustbox}
\end{table*}

% \newpage
% \quad

% \newpage

\begin{table*}
\centering\small
\begin{adjustbox}{angle=270}

        \begin{tabular}{lp{7.5cm}p{7.5cm}llllll}
            \toprule
            & & & & BERT- & & & SEScore & \\
            \multicolumn{1}{c}{No.} & Reference & Model Outputs & Category & Score & PRISM & COMET & (\textit{Ours}) & Human\\ \midrule
            \multicolumn{1}{c}{1} &  The Asian Future section includes three Chinese films & There are 3 Chinese-language films in the \underline{“Asian Future”} unit. & Better & 0.912 & -2.636 & \textbf{0.434} & -3.120 & -0.033\\\\
            
            \multicolumn{1}{c}{1} &  & \underline{“Future of Asia”} unit includes three Chinese films. & Worse & 0.922 & -2.201 & \textbf{0.134} & -2.996 & -2.433\\\\\midrule
            
            \multicolumn{1}{c}{2} & In order to prevent the risk of farmers losing land, various methods are adopted such as preferred stock, rent before stock, and repurchasing. & In order to prevent the risk of farmers losing \underline{their land}, in practice, methods such as preferred shares, \underline{first lease and then share} repurchase have been produced. & Better & 0.950  & -1.961 & 0.586 & -3.912 & -5.033\\\\
             
            \multicolumn{1}{c}{2} &  & In order to prevent the risk of farmers losing their \underline{land}, in practice, methods such as preferred stock, \underline{first lease, second share}, and repurchase \underline{have emerged}. & Worse & 0.955  & -1.797 & 0.652 & -3.178 & -5.667\\\\\midrule
            
            \multicolumn{1}{c}{3} & Self-driving development in China will become more competitive. & China's self-driving \underline{development competition} may become more intense. & Better & 0.933  & -2.370 & 0.714 & -1.635 & -1.000\\\\
             
            \multicolumn{1}{c}{3} &  & \underline{The competition in China's self-driving development} may become more intense. & Worse & 0.937  & -2.130 & 0.756 & -1.228 & -1.667\\\\\midrule
            
             \multicolumn{1}{c}{4} & To celebrate the 70th Anniversary of the Founding of the People's Republic of China and the 20th Anniversary of the Establishment of the Macao Special Administrative Region, the IAM will organize a special event of public guided visit on the subject of at the Food Information Station on the first floor of Youhan Hawker Building every Wednesday ... (\textit{144 more words}) & To celebrate the 70th anniversary of the founding of the People ’ s Republic of China and the 20th anniversary of the establishment of the Macao Special Administrative Region, the Urban Services Department will conduct a special food safety public guided tour at the Food Information Station on the first floor of the Youhan Hawker Building from October to December at 3: 30 pm on Wednesdays ...(\textit{113 more words}) & Better & 0.920  & -2.386 & 0.300 & -3.799 & -15.40\\\\
             
            \multicolumn{1}{c}{4} &
            & To celebrate the seventieth anniversary of the founding of the People's Republic of China and the twentieth anniversary of the founding of the Macao Special Administrative Region, the Municipal Department will, from October to December, at 3: 30 p.m. every Wednesday ...(\textit{130 more words}) & Worse & 0.925  & -2.360 & 0.316 & -3.617 & -25.00\\\\\midrule
            \bottomrule \\
            \multicolumn{9}{p{24.5cm}}{Table 8: Example sentences in 4 relative ranking pairs assigned by BERTScore(F1), PRISM, COMET, SEScore and Human. We use ``Better" and ``Wose" to indicate the model outputs with higher and lower human ratings, respectively. We include all metric outputs and human labels on the right side of the Table. This table demonstrates some examples where SEScore and human judgement disagree about the ranking. We \textbf{bold} the metric results which produce the correct pairwise rankings. We also \underline{underline} the error spans in each model output.}
        \end{tabular}
        %\caption{}
        \label{tab:badcasestudy}

\end{adjustbox}
\end{table*}

\end{document}